\documentclass[runningheads,a4paper]{llncs}

\usepackage{hyperref}
\hypersetup{
    pdftitle={Depth Extraction from Video Using Non-parametric Sampling},    
    pdfauthor={Kevin Karsch, Ce Liu, Sing Bing Kang},     
    pdfnewwindow=true,      
    colorlinks=true,       
    linkcolor=red,          
    citecolor=green,        
    filecolor=magenta,      
    urlcolor=cyan           
}
\usepackage[font=small,labelfont=bf]{caption}

\usepackage{amsmath}
\usepackage{rotating}

\newcommand{\boldhead}[1]{\vspace{0.05in}\noindent\textbf{#1.}}

\newcommand{\ea}[0]{et al.}
\newcommand{\ve}[1]{{\bf #1}}
\newcommand{\afterfigure}{\vspace{-1.2em}}
\newcommand{\comment}[1]{}

\usepackage{amssymb}
\usepackage{graphicx}

\usepackage{url}
\urldef{\mailsa}\path|{alfred.hofmann, ursula.barth, ingrid.haas, frank.holzwarth,|
\urldef{\mailsb}\path|anna.kramer, leonie.kunz, christine.reiss, nicole.sator,|
\urldef{\mailsc}\path|erika.siebert-cole, peter.strasser, lncs}@springer.com|

\begin{document}

\mainmatter  

\title{Depth Extraction from Video \\Using Non-parametric Sampling}

\titlerunning{Depth Extraction from Video Using Non-parametric Sampling}

%
%
\author{\hspace{.3cm} Kevin Karsch \hspace{2.2cm} Ce Liu \hspace{2cm} Sing Bing Kang}
\authorrunning{Kevin Karsch, Ce Liu and Sing Bing Kang}

\institute{ 
\begin{tabular}{c}University of Illinois \\ at Urbana-Champaign \end{tabular} \hspace{.8cm}
\begin{tabular}{c}Microsoft Research \\ New England \end{tabular} \hspace{.8cm}
\begin{tabular}{c}Microsoft Research \\ {} \end{tabular}
}

%
%

\toctitle{Depth Extraction from Video Using Non-parametric Sampling}
\tocauthor{Kevin Karsch, Ce Liu and Sing Bing Kang}

\maketitle

\begin{figure}[t]
\begin{minipage}{\textwidth}
\begin{minipage}{0.485\linewidth}
\begin{center}
\includegraphics[width=.22\linewidth, trim=0pt 25pt 0pt 52pt, clip=true]{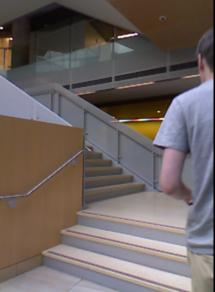}
\includegraphics[width=.22\linewidth, trim=0pt 25pt 0pt 52pt, clip=true]{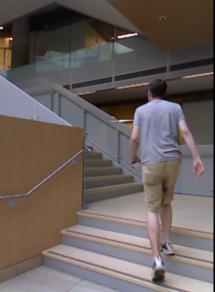}
\begin{sideways} \hspace{5mm} \tiny $\bullet$ \end{sideways}
\begin{sideways} \hspace{5mm} \tiny $\bullet$ \end{sideways}
\includegraphics[width=.22\linewidth, trim=0pt 25pt 0pt 52pt, clip=true]{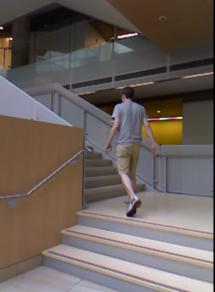}
\includegraphics[width=.22\linewidth, trim=0pt 25pt 0pt 52pt, clip=true]{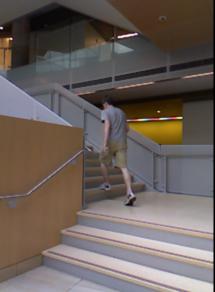}
\\
\includegraphics[width=.22\linewidth, trim=0pt 25pt 0pt 52pt, clip=true]{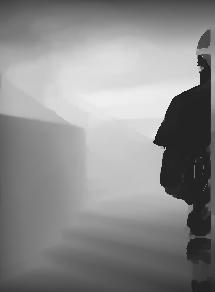}
\includegraphics[width=.22\linewidth, trim=0pt 25pt 0pt 52pt, clip=true]{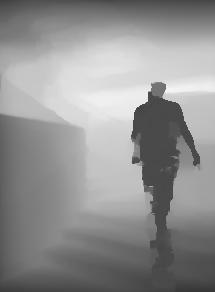}
\begin{sideways} \hspace{5mm} \tiny $\bullet$ \end{sideways}
\begin{sideways} \hspace{5mm} \tiny $\bullet$ \end{sideways}
\includegraphics[width=.22\linewidth, trim=0pt 25pt 0pt 52pt, clip=true]{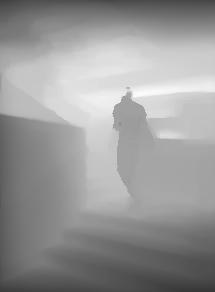}
\includegraphics[width=.22\linewidth, trim=0pt 25pt 0pt 52pt, clip=true]{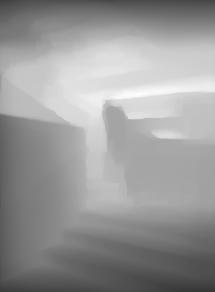}
\end{center}
\end{minipage}
\hfill
\begin{minipage}{0.485\linewidth}
\begin{center}
\includegraphics[width=.22\linewidth, trim=0pt 30pt 3pt 50pt, clip=true]{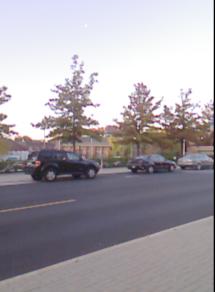}
\includegraphics[width=.22\linewidth, trim=0pt 30pt 3pt 50pt, clip=true]{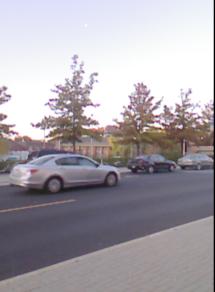}
\begin{sideways} \hspace{5mm} \tiny $\bullet$ \end{sideways}
\begin{sideways} \hspace{5mm} \tiny $\bullet$ \end{sideways}
\includegraphics[width=.22\linewidth, trim=0pt 30pt 3pt 50pt, clip=true]{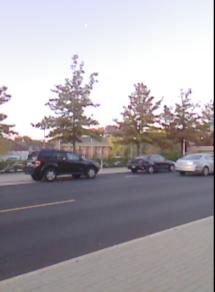}
\includegraphics[width=.22\linewidth, trim=0pt 30pt 3pt 50pt, clip=true]{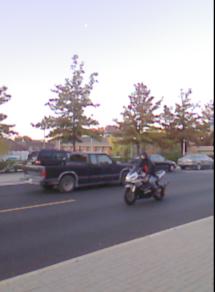}
\\
\includegraphics[width=.22\linewidth, trim=0pt 30pt 3pt 50pt, clip=true]{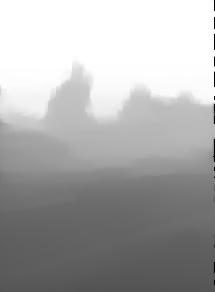}
\includegraphics[width=.22\linewidth, trim=0pt 30pt 3pt 50pt, clip=true]{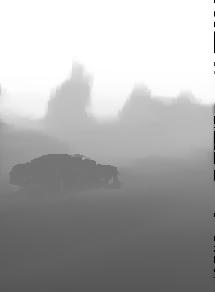}
\begin{sideways} \hspace{5mm} \tiny $\bullet$ \end{sideways}
\begin{sideways} \hspace{5mm} \tiny $\bullet$ \end{sideways}
\includegraphics[width=.22\linewidth, trim=0pt 30pt 3pt 50pt, clip=true]{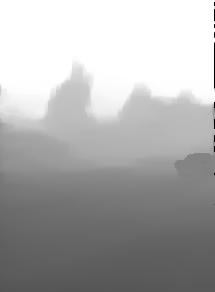}
\includegraphics[width=.22\linewidth, trim=0pt 30pt 3pt 50pt, clip=true]{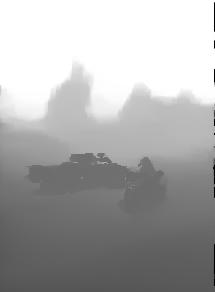}
\end{center}
\end{minipage}
\vspace{0mm}
\caption{Our technique takes a video sequence (\emph{top row}) and automatically estimates per-pixel depth (\emph{bottom row}). Our method does not require any cues from motion parallax or static scene elements; these videos were captured using a stationary camera with multiple moving objects.
\vspace{-4mm}
}
\label{fig:teaser}
\end{minipage}
\end{figure}

\begin{abstract}
\vspace{-2mm}
We describe a technique that automatically generates plausible depth maps from videos using non-parametric depth sampling. We demonstrate our technique in cases where past methods fail (non-translating cameras and dynamic scenes). Our technique is applicable to single images as well as videos. For videos, we use local motion cues to improve the inferred depth maps, while optical flow is used to ensure temporal depth consistency. For training and evaluation, we use a Kinect-based system to collect a large dataset containing stereoscopic videos with known depths. We show that our depth estimation technique outperforms the state-of-the-art on benchmark databases. Our technique can be used to automatically convert a monoscopic video into stereo for 3D visualization, and we demonstrate this through a variety of visually pleasing results for indoor and outdoor scenes, including results from the feature film {\it Charade}.
\vspace{-2mm}
\end{abstract}


\section{Introduction}
\label{sec:intro}

While many reconstruction techniques for extracting depth from video sequences exist, they typically assume moving cameras and static scenes. They do not work for dynamic scenes or for stationary, purely rotating, or strictly variable focal length sequences. There are some exceptions, e.g., \cite{Zhang:2011pami}, which can handle some moving objects, but they still require camera motion to induce parallax and allow depth estimation.

In this paper, we present a novel solution to generate depth maps from ordinary 2D videos; our solution also applies to single images. This technique is applicable to arbitrary videos, and works in cases where conventional depth recovery methods fail (static/rotating camera; change in focal length; dynamic scenes). Our primary contribution is the use of a non-parametric ``depth transfer'' approach for inferring temporally consistent depth maps without imposing requirements on the video (Sec~\ref{sec:depthest} and~\ref{sec:depthest:video}), including a method for improving the depth estimates of moving objects (Sec~\ref{sec:motion_est}). In addition, we introduce a new, ground truth stereo RGBD (RGB+depth) video dataset\footnote{Our dataset is publicly available at \url{http://kevinkarsch.com/depthtransfer}} (Sec~\ref{sec:dataset}). We also describe how we synthesize stereo videos from ordinary 2D videos using the results of our technique (Sec~\ref{sec:stereoest}).

Several reconstruction methods for single images of real, unknown scenes have also been proposed. One of the first methods, introduced by Hoiem~\ea~\cite{Hoiem:2005:APP:1186822.1073232}, created convincing reconstructions of outdoor images by assuming an image could be broken into a few planar surfaces; similarly, Delage~\ea~developed a Bayesian framework for reconstructing indoor scenes~\cite{Delage:06}. Saxena \ea~devised a supervised learning strategy for predicting depth from a single image~\cite{Saxena05learningdepth}, which was further improved to create realistic reconstructions for general scenes~\cite{Saxena:09}, and efficient learning strategies have since been proposed~\cite{Batra:CVPR:12}. Better depth estimates have been achieved by incorporating semantic labels~\cite{Liu:cvpr10}, or more sophisticated models~\cite{Li:2010}. Repetitive structures can also be used for stereo reconstruction from a single image~\cite{Wu:2011wf}. Single-image shape from shading is also possible for known (a priori) object classes~\cite{Han:2003,Hassner:06}. We not only focus on depth from a single image, but also present a framework for using temporal information for enhanced and time-coherent depth when multiple frames are available.

One application of these methods is 2D-to-3D conversion; however, many existing techniques require user interaction to refine depth and stereo estimates \cite{Guttmann:ICCV:09,Ward:11,Liao:TVCG:12}. An exception is the contemporaneous work of Konrad et al., which uses non-parametric depth sampling to automatically convert monocular images into stereoscopic images~\cite{Konrad:3DCINE:12}. Our technique extends their inference procedure, and works for videos as well.


Liu~\ea~\cite{Liu:09} showed that arbitrary scenes can be semantically labelled through non-parametric learning. Given an unlabeled input image and a database with known per-pixel labels (e.g., sky, car, tree, window), their method works by transferring the labels from the database to the input image based on SIFT features. We build on this work by transferring depth instead of semantic labels. Furthermore, we show that this ``transfer'' approach can be applied in a continuous optimization framework (Sec~\ref{sec:depthest:transfer}), whereas their method used a discrete optimization approach (MRFs). 

\section{Non-parametric depth estimation}
\label{sec:depthest}

We leverage recent work on non-parametric learning~\cite{Liu:09}, which avoids explicitly defining a parametric model and requires fewer assumptions as in past methods (e.g.,~\cite{Saxena05learningdepth,Saxena:09,Liu:cvpr10}). This approach also scales better with respect to the training data size, requiring virtually no training time. Our technique imposes no requirements on the video, such as motion parallax or sequence length, and can even be applied to a single image. We first describe our depth estimation technique as it applies to single images below, and in Sec~\ref{sec:depthest:video} we discuss novel additions that allow for improved depth estimation in videos.

\subsection{Depth estimation via continuous label transfer}
\label{sec:depthest:transfer}

\begin{figure*}[t]
\centerline{\includegraphics[width=\columnwidth]{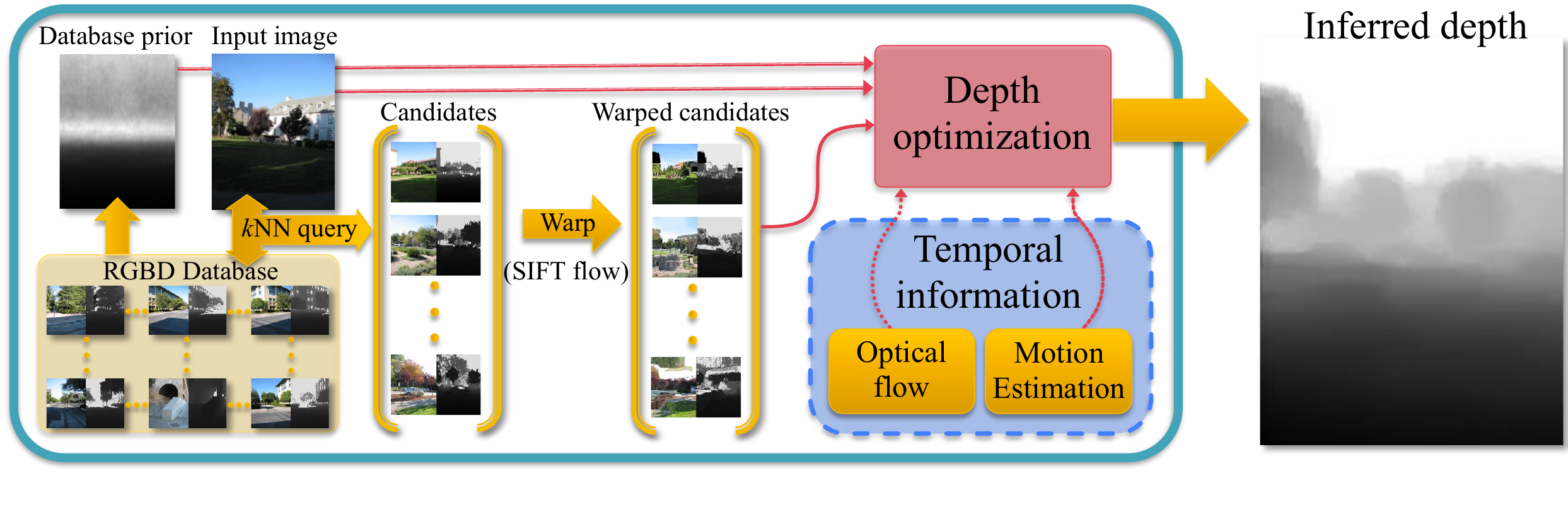}}
\vspace{-5mm}
\caption{Our pipeline for estimating depth. Given an input image, we find matching candidates in our database, and warp the candidates to match the structure of the input image. We then use a global optimization procedure to interpolate the warped candidates (Eq.~\ref{eq:depth_single}), producing per-pixel depth estimates for the input image. With temporal information (e.g. extracted from a video), our algorithm can achieve more accurate, temporally coherent depth.
}
\label{fig:depthoverview}\afterfigure
\end{figure*}

Our depth transfer approach, outlined in Fig~\ref{fig:depthoverview}, has three stages. First, given a database RGBD images, we find candidate images in the database that are ``similar'' to the input image in RGB space. Then, a warping procedure (SIFT Flow~\cite{Liu:11}) is applied to the candidate images and depths to align them with the input. Finally, an optimization procedure is used to interpolate and smooth the warped candidate depth values; this results in the inferred depth.

Our core idea is that scenes with similar semantics should have roughly similar depth distributions when densely aligned. In other words, images of semantically alike scenes are expected to have similar depth values in regions with similar appearance. Of course, not all of these estimates will be correct, which is why we find several candidate images and refine and interpolate these estimates using a global optimization technique that considers factors other than just absolute depth values.

\boldhead{RGBD database} Our system requires a database of RGBD images and/or videos.  We have collected our own RGBD video dataset, as described in Sec~\ref{sec:dataset}; a few already exist online, though they are for single images only.\footnote{Examples: Make3D range image dataset (\url{http://make3d.cs.cornell.edu/data.html}),  B3DO dataset (\url{http://kinectdata.com/}),  NYU depth datasets (\url{http://cs.nyu.edu/~silberman/datasets/}), RGB-D dataset (\url{http://www.cs.washington.edu/rgbd-dataset/}), and our own (\url{http://kevinkarsch.com/depthtransfer}).}

\boldhead{Candidate matching and warping} Given a database and an input image, we compute high-level image features (we use GIST~\cite{Oliva:ijcv01} and optical flow features, see the supplementary file) for each image or frame of video in the database as well as the input image. We then select the top $K$ ($=7$ in our work) matching frames from the database, but ensure that each video in the database contributes no more than one matching frame. This forces matching images to be from differing viewpoints, allowing for greater variety among matches. We call these matching images {\it candidate images}, and their corresponding depths {\it candidate depths}.

\begin{figure}[t]
\centerline{
\hfill
(a)\includegraphics[width=0.3\columnwidth,page=1]{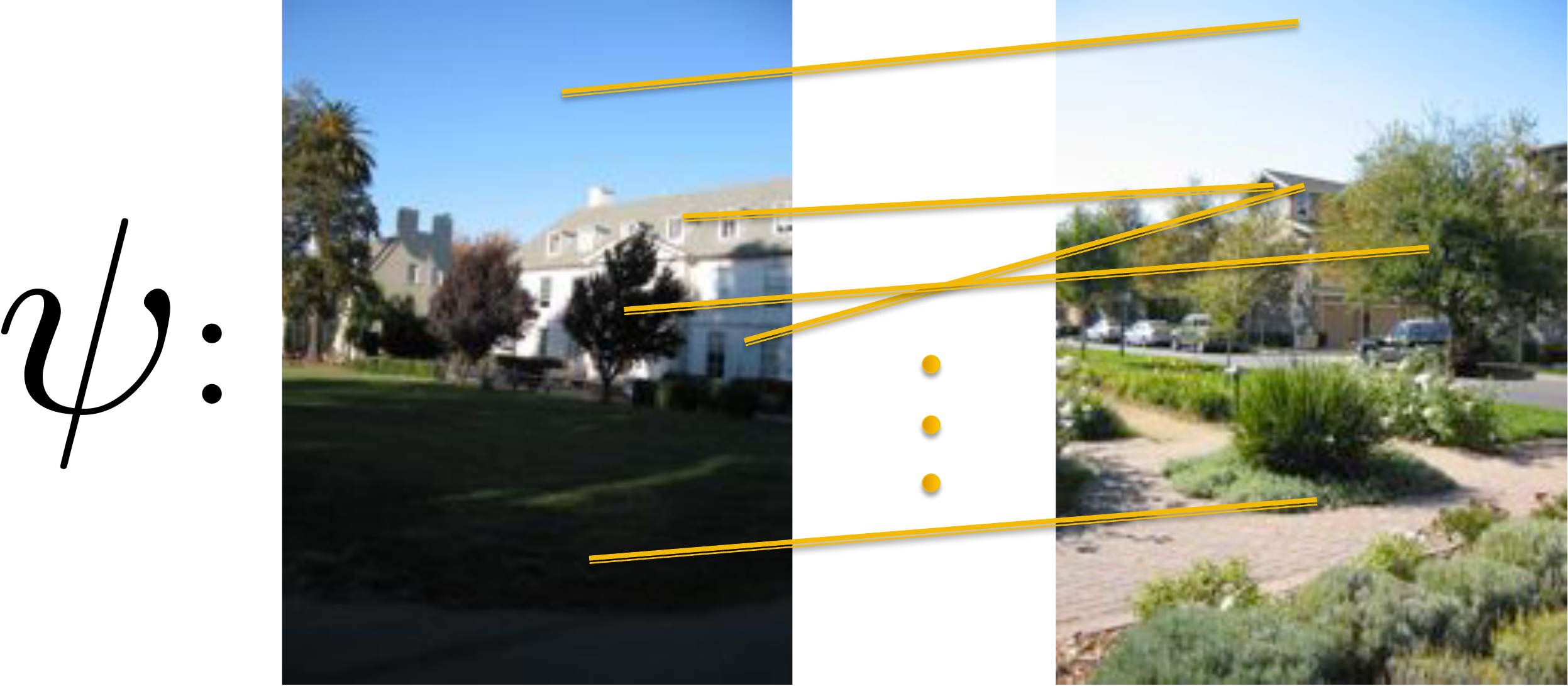}
\hfill
(b)\includegraphics[width=0.3\columnwidth,page=2]{fig/siftflow}
\hfill
}
\caption{SIFT flow warping. (a) SIFT features are calculated and matched in a one-to-many fashion, which defines $\psi$. (b) $\psi$ is applied to achieve dense scene alignment.
}
\label{fig:siftflow}\afterfigure
\end{figure}

Because the candidate images match the input closely in feature space, it is expected that the overall semantics of the scene are roughly similar. We also make the critical assumption that the distribution of depth is comparable among the input and candidates. However, we want pixel-to-pixel correspondences between the input and all candidates, as to limit the search space when inferring depth from the candidates.

We achieve this pixel-to-pixel correspondence through SIFT flow~\cite{Liu:11}, which matches per-pixel SIFT features to estimate dense scene alignment. Using SIFT flow, we estimate warping functions $\psi_i, i \in \{1,\ldots,K\}$ for each candidate image; this process is illustrated in Fig~\ref{fig:siftflow}. These warping functions map pixel locations from a given candidate's domain to pixel locations in the input's domain. The warping functions can be one-to-many.

\boldhead{Depth optimization}
Each warped candidate depth is deemed to be a rough approximation of the input's depth map. Unfortunately, such candidate depths may still contain inaccuracies and are often not spatially smooth. Instead, we generate the most likely depth map by considering all of the warped candidates, optimizing with spatial regularization in mind.

Let $\ve{L}$ be the input image and $\ve{D}$ the depth map we wish to infer. We minimize
\begin{equation}
\label{eq:depth_single}
-\log( P(\ve{D} | \ve{L}) ) = E(\ve{D}) =  \sum_{i\in\text{pixels}} E_{\text{t}}(\ve{D}_i) + \alpha E_{\text{s}}(\ve{D}_i) + \beta E_{\text{p}}(\ve{D}_i) + \log(Z),
\end{equation}
where $Z$ is the normalization constant of the probability, and $\alpha$ and $\beta$ are parameters ($\alpha=10, \beta=0.5$). For a single image, our objective contains three terms: data ($E_{\text{t}}$), spatial smoothness ($E_{\text{s}}$), and database prior ($E_{\text{p}}$). 

The data term measures how close the inferred depth map $\ve{D}$ is to each of the warped candidate depths, $\psi_j(C^{(j)})$. This distance measure is defined by $\phi$, a robust error norm (we use an approximation to the $L1$ norm, $\phi(x) = \sqrt{x^2+\epsilon}$, with $\epsilon = \text{10}^{-4}$). We define the data term as\vspace{-.5em}
\begin{equation}
\label{eq:transfer}
\begin{split}
E_{\text{t}}(\ve{D}_i) =  \sum_{j=1}^K w^{(j)}_i \Bigl[ & \phi(\ve{D}_i-\psi_j(C^{(j)}_i)) +  \\ 
& \gamma \bigl[ \phi(\nabla_x  \ve{D}_i-\psi_j(\nabla_x C^{(j)}_i)) + \phi(\nabla_y \ve{D}_i-\psi_j(\nabla_y C^{(j)}_i)) \bigr]  \Bigr],
\end{split}
\end{equation}\vspace{-1.2em}\\
where $w_i^{(j)}$ is a confidence measure of the accuracy of the $j^{th}$ candidate's warped depth at pixel $i$ (more details in the supplementary file), and $K$ ($=7$) is the total number of candidates. We measure not only absolute differences, but also relative depth changes, i.e., depth gradients. The latter terms of Eq~\ref{eq:transfer} enforce similarity among candidate depth gradients and inferred depth gradients, weighted by $\gamma$ ($=10$).

We encourage spatial smoothness, but more so in regions where the input image has small intensity gradients:\vspace{-.5em}
\begin{equation}
\label{eq:smooth}
E_{\text{s}}(\ve{D}_i) = s_{x,i} \phi(\nabla_x \ve{D}_i) + s_{y,i} \phi(\nabla_y \ve{D}_i).
\end{equation}
The depth gradients along x and y ($\nabla_x \ve{D}, \nabla_y \ve{D}$) are modulated by soft thresholds (sigmoidal functions) of image gradients in the same directions ($\nabla_x \ve{L}, \nabla_y \ve{L}$), namely, $s_{x,i} = (1+e^{(||\nabla_x \ve{L}_i||-0.05)/.01})^{-1}$ and $s_{y,i} = (1+e^{(||\nabla_y \ve{L}_i||-0.05)/.01})^{-1}$; see the supplemental file for further explanation.

We also incorporate assumptions from our database that will guide the inference when pixels have little or no influence from other terms (due to weights $w$ and $s$):
\begin{equation}
\label{eq:prior}
E_{\text{p}}(\ve{D}_i)= \phi(\ve{D}_i - \mathcal{P}_i).
\end{equation}
We compute the prior, $\mathcal{P}$, by averaging all depth images in our database.

This is an unconstrained, non-linear optimization, and we use iteratively reweighted least squares to minimize our objective function (details in the supplementary file).

\section{Improving depth estimation for videos}
\label{sec:depthest:video}

Generating depth maps frame-by-frame without incorporating temporal information often leads to temporal discontinuities; past methods that ensure temporal coherence rely on a translating camera and static scene objects.  Here, we present a framework that improves depth estimates and enforces temporal coherence for {\it arbitrary video sequences}. That is, our algorithm is suitable for videos with moving scene objects and rotating/zooming views where conventional SFM and stereo techniques fail. (Here, we assume that zooming induces little or no parallax.)

\begin{figure}[t]

\begin{minipage}{0.42\linewidth}
\begin{center}
\begin{minipage}{.23\linewidth}
\includegraphics[width=\linewidth, trim=0pt 30pt 1pt 48pt, clip=true]{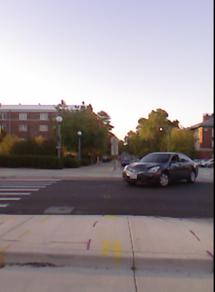}
\includegraphics[width=\linewidth, trim=0pt 30pt 1pt 48pt, clip=true]{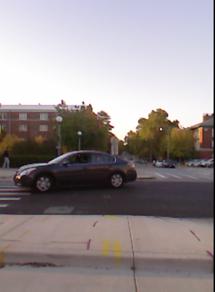}
\includegraphics[width=\linewidth, trim=0pt 30pt 1pt 48pt, clip=true]{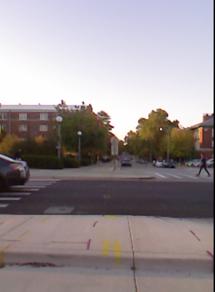}
\end{minipage}
\begin{minipage}{.23\linewidth}
\includegraphics[width=\linewidth, trim=0pt 30pt 1pt 48pt, clip=true]{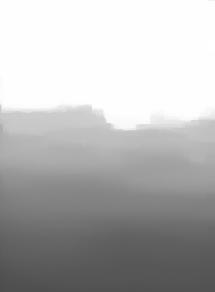}
\includegraphics[width=\linewidth, trim=0pt 30pt 1pt 48pt, clip=true]{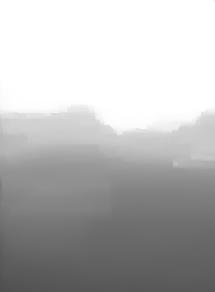}
\includegraphics[width=\linewidth, trim=0pt 30pt 1pt 48pt, clip=true]{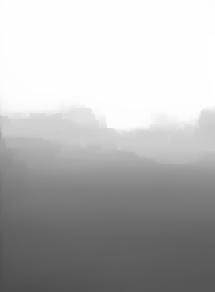}
\end{minipage}
\begin{minipage}{.23\linewidth}
\vspace{.5mm}
\includegraphics[width=\linewidth, trim=0pt 30pt 1pt 48pt, clip=true]{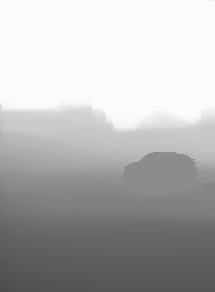}
\includegraphics[width=\linewidth, trim=0pt 30pt 1pt 48pt, clip=true]{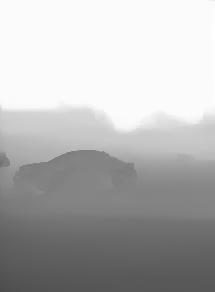}
\includegraphics[width=\linewidth, trim=0pt 30pt 1pt 48pt, clip=true]{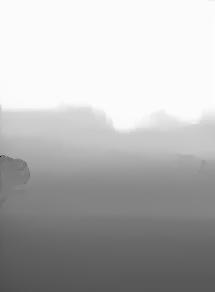}
\end{minipage}
\begin{minipage}{.23\linewidth}
\includegraphics[width=\linewidth, trim=0pt 30pt 1pt 48pt, clip=true]{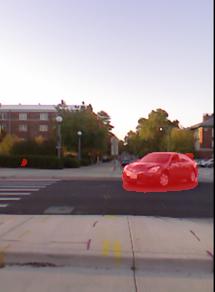}
\includegraphics[width=\linewidth, trim=0pt 30pt 1pt 48pt, clip=true]{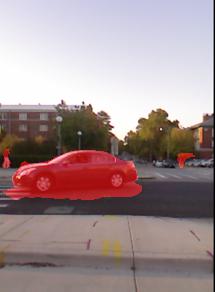}
\includegraphics[width=\linewidth, trim=0pt 30pt 1pt 48pt, clip=true]{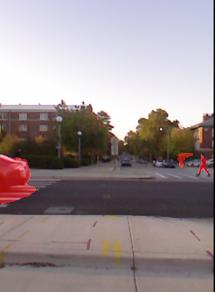}
\end{minipage}
\end{center}
\vspace{-5mm}
\caption{Importance of temporal information. Left: input frames. Mid-left: predicted depth without temporal information. Note that the car is practically ignored here. Mid-right: predicted depth with temporal information, with the depth of the moving car recovered. Right: detected moving object.
}\afterfigure
\label{fig:temporal_benefit}
\end{minipage}
\hfill
\begin{minipage}{0.55\linewidth}
\begin{center}
\includegraphics[width=1\linewidth,page=1]{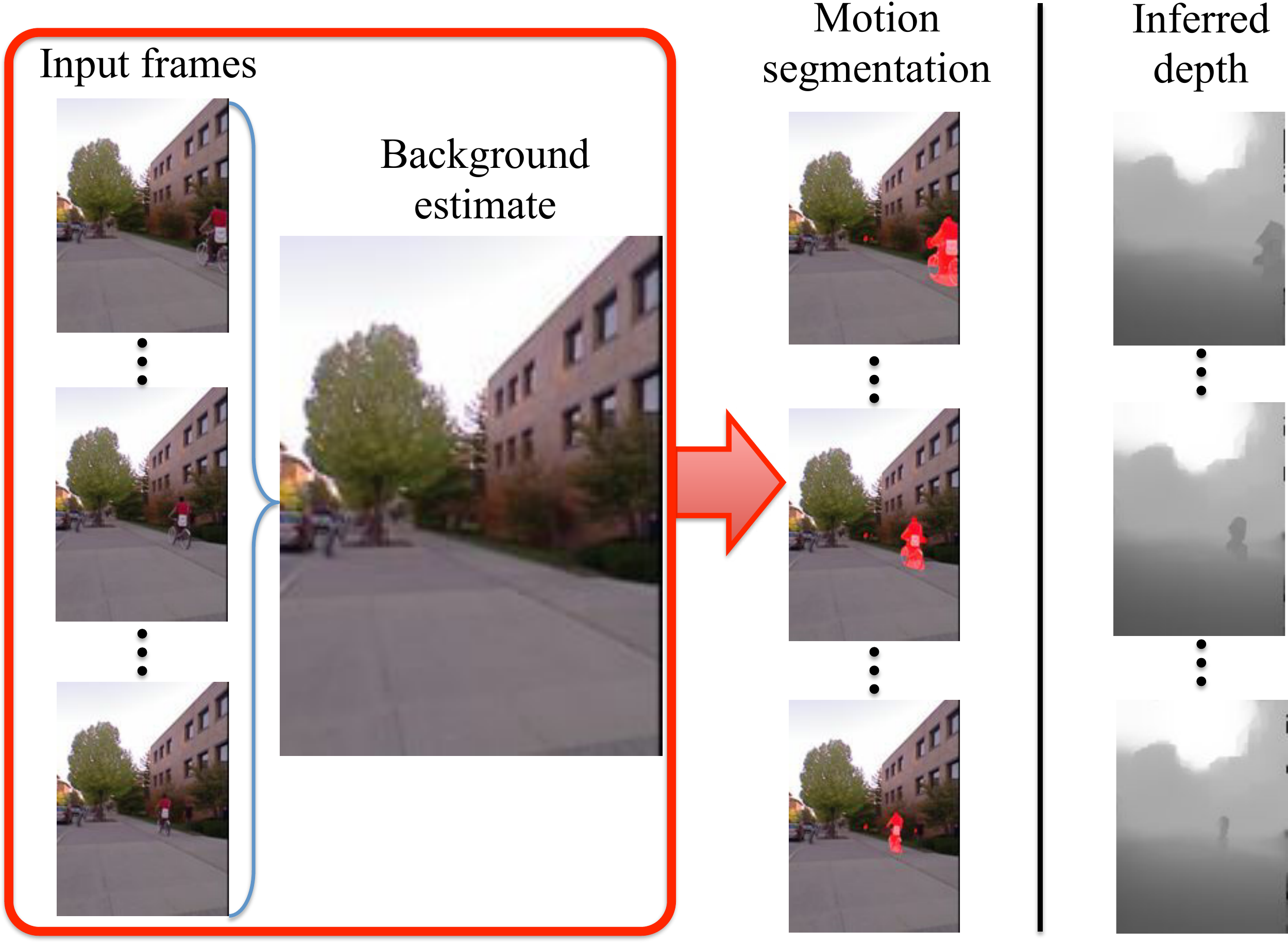}
\end{center}
\vspace{-5mm}
\caption{We apply median filtering on the stabilized images to extract the background image. A pixel is deemed to be in motion if there is sufficient relative difference from the background image.
\vspace{2mm}
}
\label{fig:motionseg}\afterfigure
\end{minipage}

\end{figure}

Our idea is to incorporate temporal information through additional terms in the optimization that ensure (a) depth estimates are consistent over time and (b) that moving objects have depth similar to their contact point with the ground. Each frame is processed the same as in the single image case (candidate matching and warping), except that now we employ a global optimization (described below) that infers depth for the entire sequence at once, incorporating temporal information from all frames in the video.
Fig~\ref{fig:temporal_benefit} illustrates the importance of these additional terms in our optimization.

We formulate the objective for handling video by adding two terms to the single-image objective:
\begin{equation}
\label{eq:depth_video}
E_{\text{video}}(\ve{D}) = E(\ve{D}) +  \sum_{i\in\text{pixels}} \nu  E_{\text{c}}(\ve{D}_i) + \eta E_{\text{m}}(\ve{D}_i),
\end{equation}
where $E_{\text{c}}$ encourages temporal coherence while $E_{\text{m}}$ uses motion cues to improve the depth of moving objects. The weights $\nu$ and $\eta$ balance the relative influence of each term ($\nu = 100, \eta = 5$).

We model temporal coherence first by computing per-pixel optical flow for each pair of consecutive frames in the video (using Liu's publicly available code~\cite{Liu:thesis09}). We define the {\it flow difference}, $\nabla_{flow}$, as a linear operator which returns the change in the flow across two corresponding pixels, and model the coherence term as
\begin{equation}
\label{eq:coherence}
E_{\text{c}}(\ve{D}_i) =   s_{t,i} \phi(\nabla_{flow}\ve{D}_i).
\end{equation}
We weight each term by a measure of flow confidence, \\$s_{t,i} = (1+e^{-(||\nabla_{flow} \ve{L}_i||-0.05)/.01})^{-1}$, which intuitively is a soft threshold on the reprojection error. Minimizing the weighted flow differences has the effect of temporally smoothing inferred depth in regions where optical flow estimates are accurate.

To handle motion, we detect moving objects in the video (Sec~\ref{sec:motion_est}) and constrain their depth such that these objects touch the floor. Let $m$ be the binary motion segmentation mask and $\mathcal{M}$ the depth in which connected components in the segmentation mask contact the floor. We define the motion term as
\begin{equation}
\label{eq:motion}
E_{\text{m}}(\ve{D}_i) = m_i \phi(\ve{D}_i - \mathcal{M}_i).
\end{equation}

\subsection{Motion segmentation}
\label{sec:motion_est}

Differentiating moving and stationary objects in the scene can be a useful cue when estimating depth. Here we describe our algorithm for detecting objects in motion in non-translating movies (i.e., static, rotational, and variable focal length videos).

First, to account for dynamic exposure changes throughout the video, we find the image with the lowest overall intensity in the sequence and perform histogram equalization on all other frames in the video. We use this image as to not propagate spurious noise found in brighter images. Next, we use RANSAC on point correspondences to compute the dominant camera motion (modeled using homography) to align neighboring frames in the video.  Median filtering is then used on the stabilized images to extract the background $B$ (ideally, without all the moving objects). 


In our method, the likelihood of a pixel being in motion depends on how different it is from the background, weighted by the optical flow magnitude which is computed between stabilized frames (rather than between the original frames). We use relative differencing (relative to background pixels) to reduce reliance on absolute intensity values, and then threshold to produce a mask:\vspace{-.5em}
\begin{equation}
\label{eq:moseg}
m_{i,k} = \left\{ \begin{array}{ll} 1 & \text{if } ||flow_{i,k}|| \frac{||W_{i,k}-B_i||^2}{B_i} > \tau \\ 0 & \text{otherwise,} \end{array} \right.
\end{equation}\vspace{-1.em}\\
where $\tau = 0.01$ is the threshold, and $W_{i,k}$ is the $k^{th}$ pixel of the $i^{th}$ stabilized frame (i.e., warped according to the homography that aligns $W$ with $B$). This produces a motion estimate in the background's coordinate system, so we apply the corresponding inverse homography to each warped frame to find the motion relative to each frame of the video. This segmentation mask is used (as in Eq~\ref{eq:motion}) to improve depth estimates for moving objects in our optimization. Fig~\ref{fig:motionseg} illustrates this technique.

\section{Dataset}
\label{sec:dataset}

\begin{figure}[t]
\centerline{\includegraphics[width=.65\columnwidth]{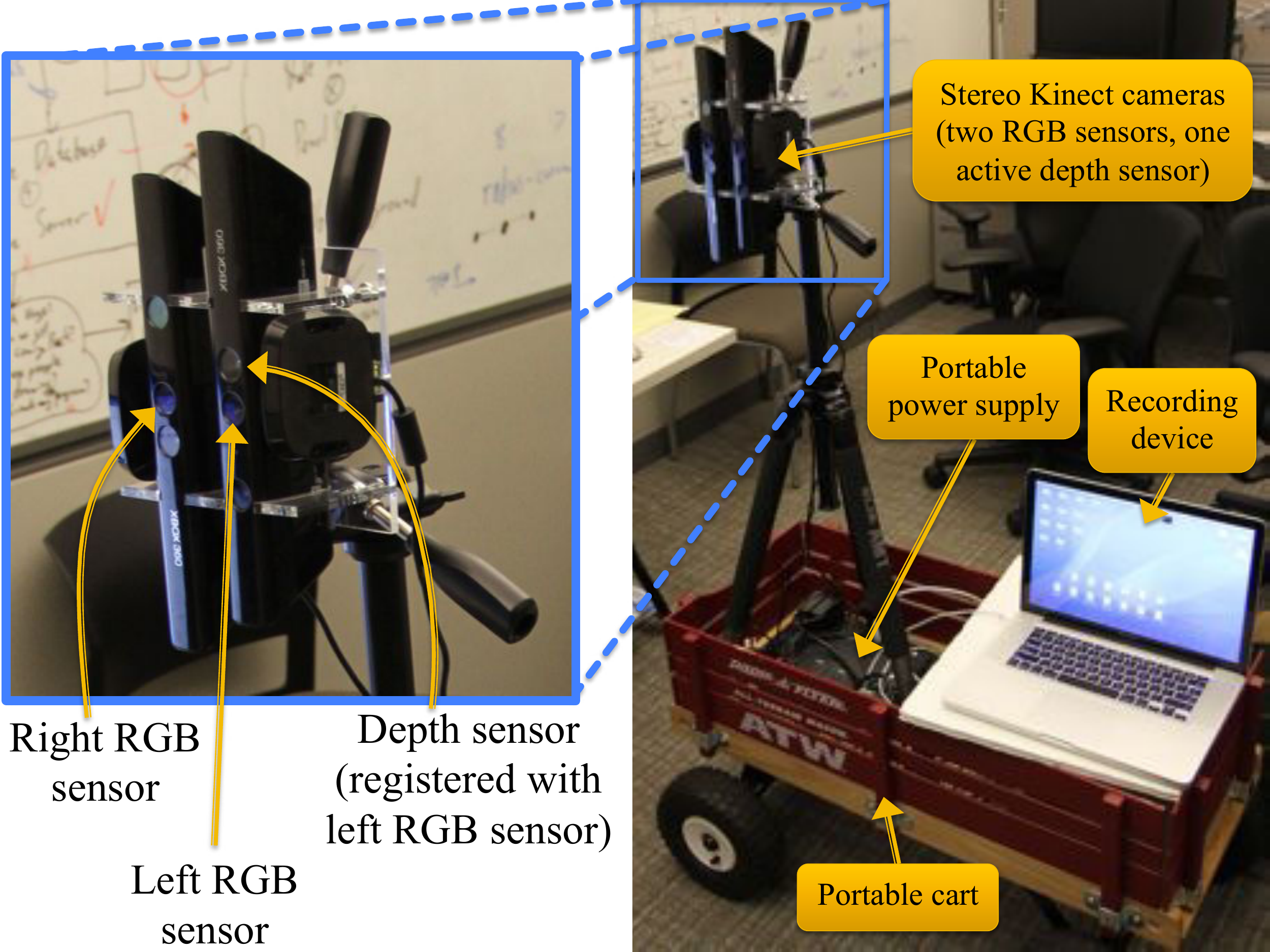}}
\caption{Our stereo-RGBD collection rig consists of two side-by-side Microsoft Kinects. The rig is mobile through the use of an uninterruptible power supply, laptop, and rolling mount.
}\afterfigure
\label{fig:kinectrig}
\end{figure}

In order to train and test our technique on video input, we collected a dataset of over 200 stereoscopic video sequences with corresponding depth values for one of the two stereoscopic views. These sequences come from four different buildings in two cities and contain substantial scene variation (e.g., hallways, rooms, foyers). Each clip is filmed with camera viewpoints that are either static or slowly rotated. Our dataset primarily contains one or more persons walking through a scene, sitting or socializing.

To capture data, we use two side-by-side, vertically mounted Microsoft Kinects shown in Fig~\ref{fig:kinectrig} (positioned about 5cm apart). We collected the color images from both Kinects and only the depth map from the left Kinect.

We also collected outdoor data with our stereo device. However, because the Kinect cannot produce depth maps outdoors due to IR interference from the sunlight, we could not use these sequences for training. We did not apply stereo to extract ground truth depth because of reliability issues. We did, however, use this data for testing and evaluation purposes.

\begin{figure*}[t]
\begin{center}
\begin{minipage}{0.49\linewidth}
\includegraphics[width=.32\linewidth]{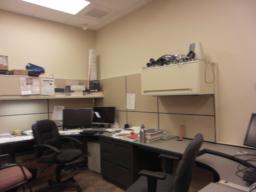}
\includegraphics[width=.32\linewidth]{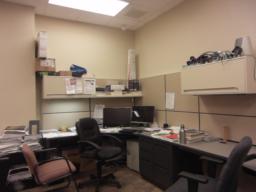}
\includegraphics[width=.32\linewidth]{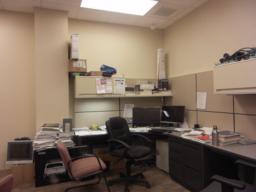}
\\
\includegraphics[width=.32\linewidth]{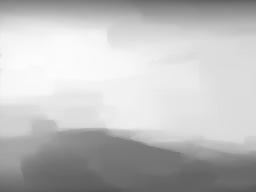}
\includegraphics[width=.32\linewidth]{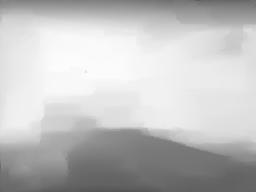}
\includegraphics[width=.32\linewidth]{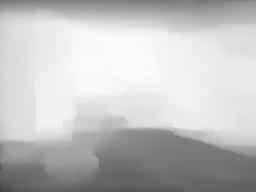}
\\
\includegraphics[width=.32\linewidth]{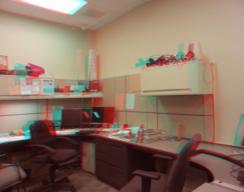}
\includegraphics[width=.32\linewidth]{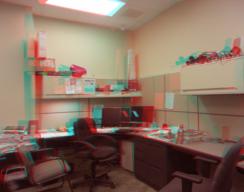}
\includegraphics[width=.32\linewidth]{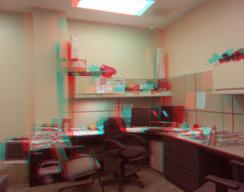}
\end{minipage}
\hfill
\rule[-22.5mm]{.15mm}{.375\linewidth}
\hfill
\begin{minipage}{0.49\linewidth}
\includegraphics[width=.32\linewidth]{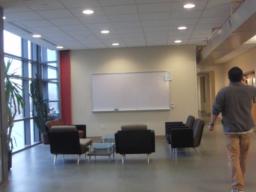}
\includegraphics[width=.32\linewidth]{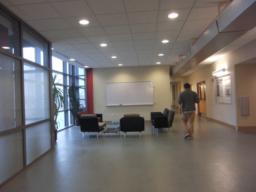}
\includegraphics[width=.32\linewidth]{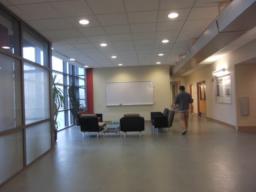}
\\
\includegraphics[width=.32\linewidth]{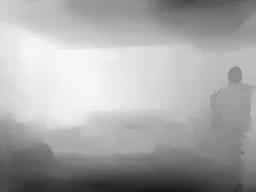}
\includegraphics[width=.32\linewidth]{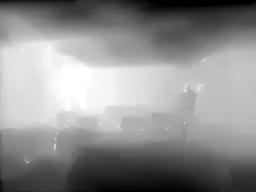}
\includegraphics[width=.32\linewidth]{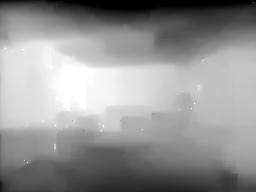}
\\
\includegraphics[width=.32\linewidth]{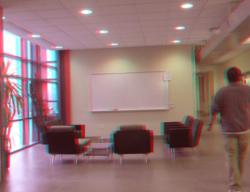}
\includegraphics[width=.32\linewidth]{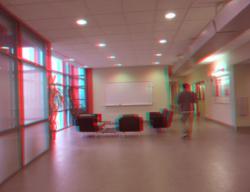}
\includegraphics[width=.32\linewidth]{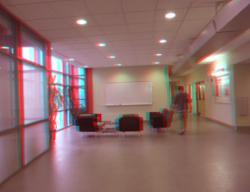}
\end{minipage}
\\ 
\rule{\linewidth}{.15mm}
\\ 
\begin{minipage}{0.49\linewidth}
\includegraphics[width=.32\linewidth]{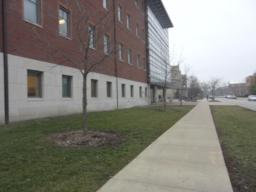}
\includegraphics[width=.32\linewidth]{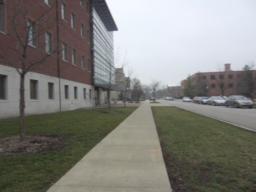}
\includegraphics[width=.32\linewidth]{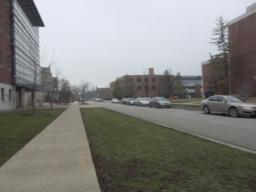}
\\
\includegraphics[width=.32\linewidth]{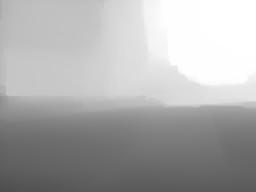}
\includegraphics[width=.32\linewidth]{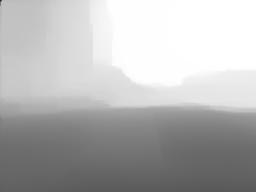}
\includegraphics[width=.32\linewidth]{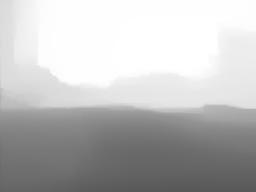}
\\
\includegraphics[width=.32\linewidth]{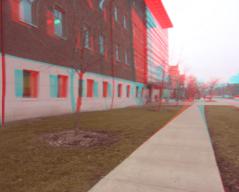}
\includegraphics[width=.32\linewidth]{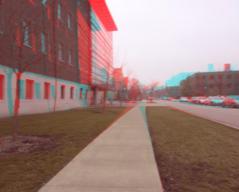}
\includegraphics[width=.32\linewidth]{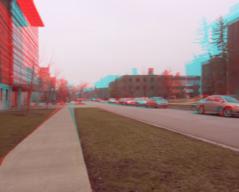}
\end{minipage}
\hfill
\rule[-22mm]{.15mm}{.38\linewidth}
\hfill
\begin{minipage}{0.49\linewidth}
\includegraphics[width=.32\linewidth]{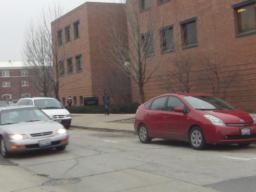}
\includegraphics[width=.32\linewidth]{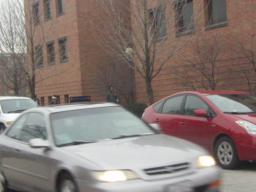}
\includegraphics[width=.32\linewidth]{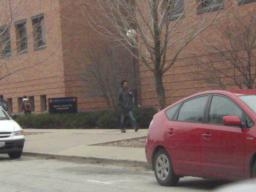}
\\
\includegraphics[width=.32\linewidth]{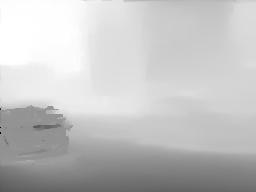}
\includegraphics[width=.32\linewidth]{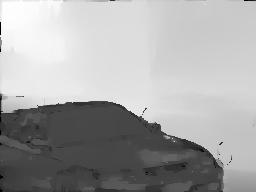}
\includegraphics[width=.32\linewidth]{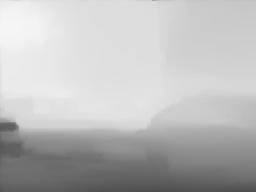}
\\
\includegraphics[width=.32\linewidth]{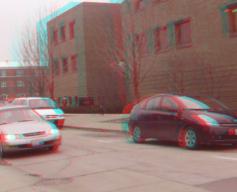}
\includegraphics[width=.32\linewidth]{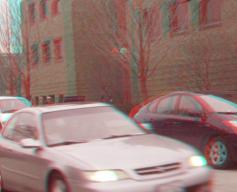}
\includegraphics[width=.32\linewidth]{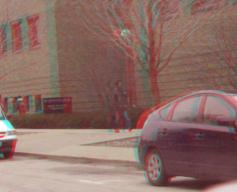}
\end{minipage}
\end{center}
\vspace{-5mm}
\caption{Results obtained on four different sequences sequences captured with a rotating camera and/or variable focal length. In each 3$\times$3 block of images, we show the input frames (\emph{top}), inferred depth (\emph{middle}) and inferred 3D anaglyph (\emph{bottom}). Notice that the sequences are time-coherent and that moving objects are not ignored.
\vspace{-5mm}
}
\label{fig:rot_zoom_results}
\end{figure*}

\section{Application: Automatic stereoscopic view synthesis}
\label{sec:stereoest}

In recent years, 3D\footnote{The presentation of stereoscopic (left+right) video to convey the sense of depth.} videos have become increasingly popular. Many feature films are now available in 3D, and increasingly more personal photography devices are now equipped with stereo capabilities (from point-and-shoots to attachments for video cameras and SLRs). Distributing user-generated content is also becoming easier. Youtube has recently incorporated 3D viewing and uploading features, and many software packages have utilities for handling and viewing 3D file formats, e.g., Fujifilm's FinePixViewer.

As 3D movies and 3D viewing technology become more widespread, 
it is desirable to have techniques that can convert legacy 2D movies to 3D in an efficient and inexpensive way. Currently, the movie industry uses expensive solutions that tend to be manual-intensive. For example, it was reported that the cost of converting (at most) 20 minutes of footage for the movie ``Superman Returns'' was \$10 million\footnote{See \url{http://en.wikipedia.org/wiki/Superman\_Returns}.}.

Our technique can be used to automatically generate the depth maps necessary to produce the stereoscopic video (by warping each input frame using its corresponding depth map). To avoid generating holes at disocclusions in the view synthesis step, we adapt and extend Wang~\ea's technique~\cite{Wang:sbim11}. They developed a method that takes as input a single image and per-pixel disparity values, and intelligently warps the input image based on the disparity such that highly salient regions remain unmodified. Their method was applied only to single images; we extend this method to handle video sequences as well. Details of our view synthesis technique are given in the supplementary file.

\section{Results}
\vspace{-0.03mm}
\label{sec:results}


\begin{figure*}[t]
\begin{minipage}{0.49\linewidth}
\includegraphics[width=.24\linewidth]{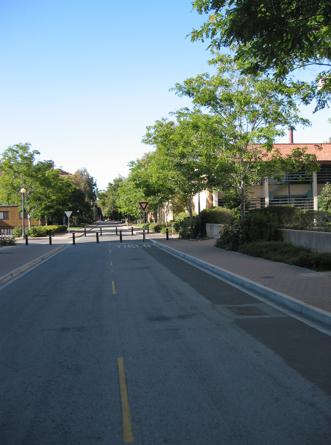}
\includegraphics[width=.24\linewidth]{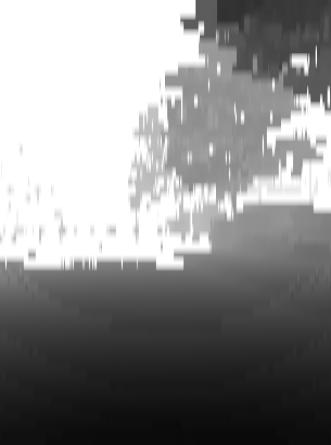}
\includegraphics[width=.24\linewidth]{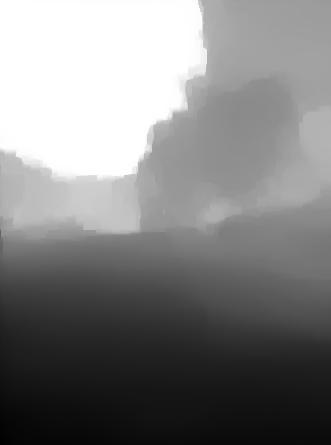}
\includegraphics[height=.323\linewidth]{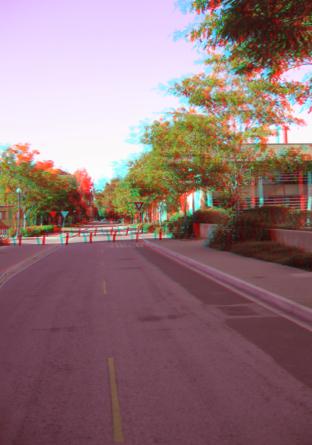}
\end{minipage}
\hfill
\rule[-9.5mm]{.15mm}{.165\linewidth}
\hfill
\begin{minipage}{0.49\linewidth}
\includegraphics[width=.235\linewidth]{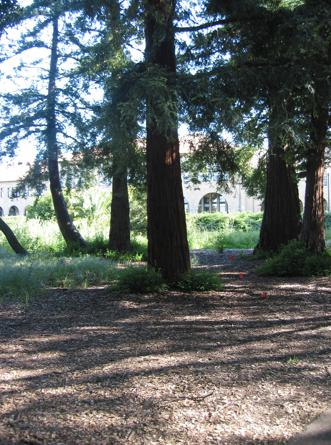}
\includegraphics[width=.235\linewidth]{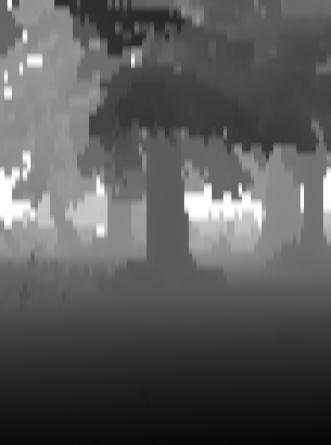}
\includegraphics[width=.235\linewidth]{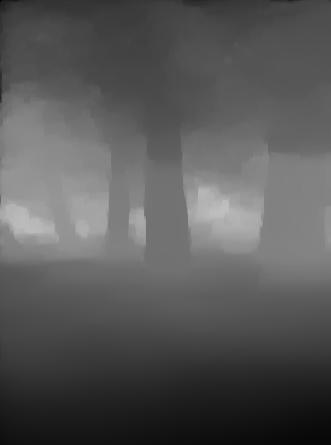}
\includegraphics[height=.315\linewidth]{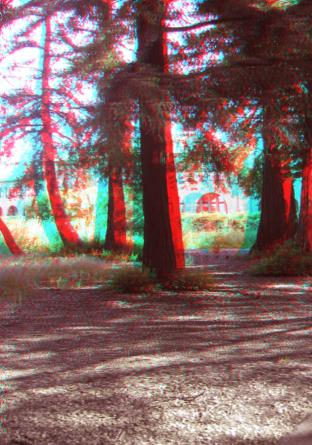}
\end{minipage}
\\
\rule{.995\linewidth}{.15mm}
\\
\begin{minipage}{0.49\linewidth}
\includegraphics[width=.24\linewidth]{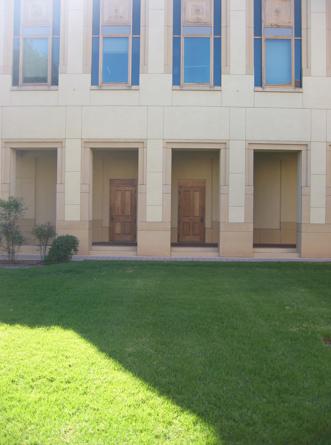}
\includegraphics[width=.24\linewidth]{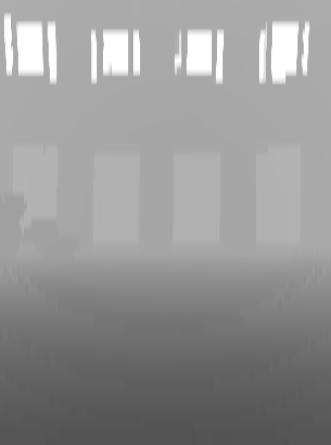}
\includegraphics[width=.24\linewidth]{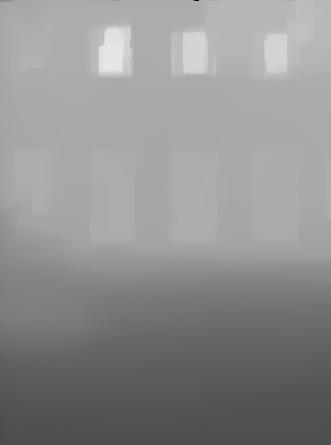}
\includegraphics[height=.3225\linewidth]{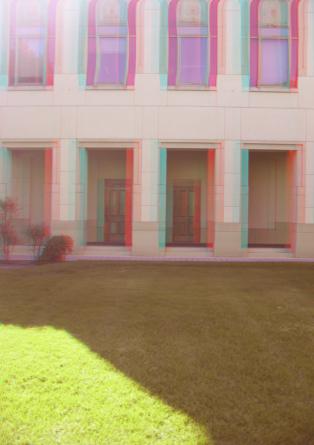}
\end{minipage}
\hfill
\rule[-9mm]{.15mm}{.165\linewidth}
\hfill
\begin{minipage}{0.49\linewidth}
\includegraphics[width=.235\linewidth]{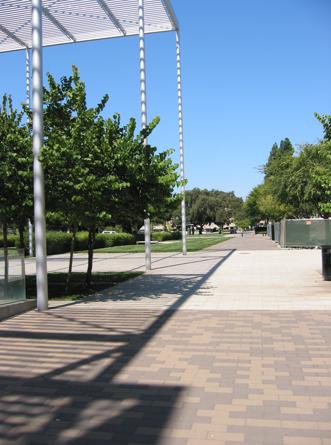}
\includegraphics[width=.235\linewidth]{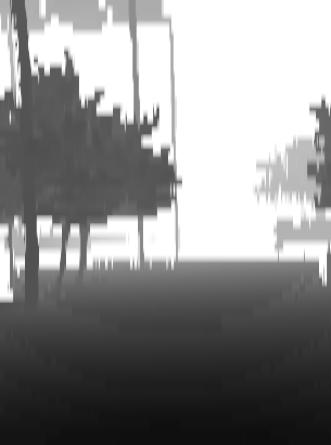}
\includegraphics[width=.235\linewidth]{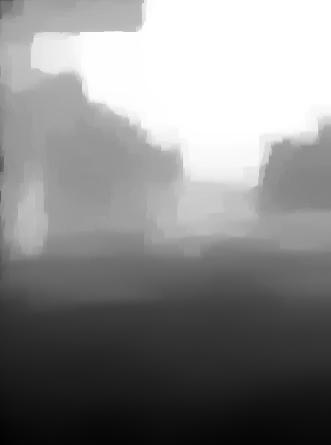}
\includegraphics[height=.32\linewidth]{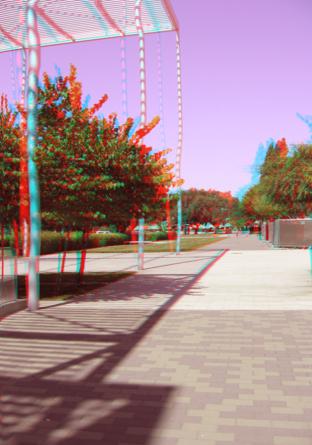}
\end{minipage}
\vspace{-4mm}
\caption{Single image results obtained on test images from the Make3D dataset. Each result contains the following four images (from left to right): original photograph,  ground truth depth from the dataset, our inferred depth, and our synthesized anaglyph \protect\includegraphics[height=5pt]{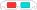} image. The depth images are shown in log scale. Darker pixels indicate nearby objects (black is roughly 1m away) and lighter pixels indicate objects farther away (white is roughly 80m away). Each pair of ground truth and inferred depths are displayed at the same scale.
}
\label{fig:outdoor_results}\afterfigure
\end{figure*}

We use the Make3D range image dataset to evaluate our single image depth estimation technique. Of the 534 images in the Make3D dataset, we use 400 for testing and 134 for training (the same as was done before, e.g.,~\cite{Saxena05learningdepth,Saxena:09,Li:2010,Liu:cvpr10}). We report error for three common metrics in Table~\ref{tab:results}. Denoting $\ve{D}$ as estimated depth and $\ve{D}^*$ as ground truth depth, we compute {\it relative} ({\bf rel}) {\it error} $\frac{|\ve{D}-\ve{D}^*|}{\ve{D}^*}$, {\bf log$_{10}$} {\it error} $|\log_{10}(\ve{D})-\log_{10}(\ve{D}^*)|$, and {\it root mean squared} ({\bf RMS}) {\it error} $\sqrt{ \sum_{i=1}^N (\ve{D}_i-\ve{D}_i^*)^2/N}$. Error measures are averaged over all pixels/images in the test set. Our estimated depth maps are computed at 345$\times$460 pixels (maintaining the aspect ratio of the Make3D dataset input images).

Our method is as good as or better than the state-of-the-art for each metric. Note that previously no method achieved state-of-the-art results in more than one metric. Our estimated depth is good enough to generate compelling 3D images, and representative results are shown in Fig~\ref{fig:outdoor_results}. In some cases, our method even produces better depth estimates than the ground truth (with low resolution and sensor errors). Thin structures (e.g., trees and pillars) are usually recovered well; however, fine structures are occasionally missed due to spatial regularization (such as the poles in the bottom-right image of Fig~\ref{fig:outdoor_results}).

\begin{table}[t]
\begin{minipage}{0.5\linewidth}
\begin{tabular}{| l || c | c | c |} \hline
{\bf Method} & {\bf rel }& {\bf log$_{10}$} & {\bf RMS} \\ \hline \hline
 Depth MRF~\cite{Saxena05learningdepth} & 0.530 & 0.198 & 16.7 \\ \hline
 Make3D~\cite{Saxena:09} & 0.370 & 0.187 & N/A \\ \hline
 Feedback Cascades~\cite{Li:2010} & N/A & N/A & 15.2 \\ \hline
 Semantic Labels~\cite{Liu:cvpr10} & 0.375 & {\bf 0.148} & N/A \\ \hline
 Depth Transfer (ours) & {\bf 0.361} & {\bf 0.148} & {\bf 15.1} \\ \hline
 \end{tabular}
\vspace{2mm}
\caption{Comparison of depth estimation errors on the Make3D range image dataset. Using our single image technique, our method achieves state of the art results in each metric ({\bf rel} is relative error,  {\bf RMS} is root mean squared error; details in text).}
 \label{tab:results}
\end{minipage}
\hfill
\begin{minipage}{0.46\linewidth}
\vspace{-4mm}
\begin{tabular}{| l || c | c | c || c |} \hline
{\bf Dataset} & {\bf rel }& {\bf log$_{10}$} & {\bf RMS} & {\bf PSNR}\\ \hline \hline
 Building 1$^{\dagger}$ & 0.196 & 0.082 & 8.271 & 15.6 \\ \hline
 Building 2 & 0.394 & 0.135 & 11.7 & 15.6 \\ \hline
 Building 3 & 0.325 & 0.159 & 15.0 & 15.0\\ \hline
 Building 4 & 0.251 & 0.136 & 15.3 & 16.4 \\ \hline
 Outdoors$^{\dagger\dagger}$ & N/A & N/A & N/A & 15.2 \\ \hline \hline
 All & 0.291 & 0.128 & 12.6 & 15.6 \\ \hline
 \end{tabular}
\vspace{0.5mm}
\caption{Error averaged over our stereo-RGBD dataset. 
${}^{\dagger}$Building used for training (results for Building 1 trained using a hold-one-out scheme). $^{\dagger\dagger}$No ground truth depth available.
}
 \label{tab:results_kinect}\afterfigure
\end{minipage}
\vspace{-7mm}
\end{table}

\begin{figure}
\begin{center}
\begin{minipage}{.8\linewidth}
\hfill
\begin{minipage}{.17\linewidth}
\centerline{ {\bf Input}}
\centerline{215$\times$292}
\includegraphics[width=\linewidth]{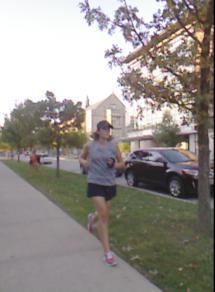}
\includegraphics[width=\linewidth]{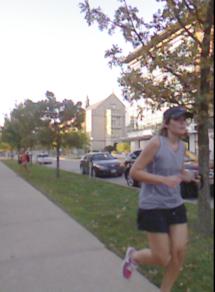}
\end{minipage}
\hfill
\begin{minipage}{.34\linewidth}
\centerline{ {\bf Make3D}}
\centerline{305$\times$55}
\begin{minipage}{.5\linewidth}
\includegraphics[width=\linewidth]{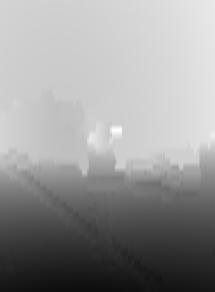}
\includegraphics[width=\linewidth]{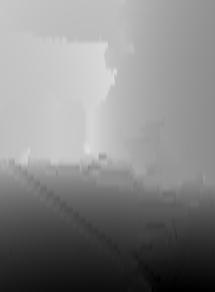}
\end{minipage}
\begin{minipage}{.465\linewidth}
\includegraphics[width=\linewidth]{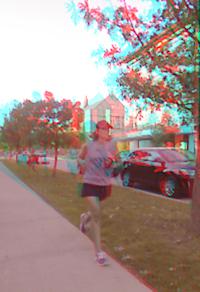}
\includegraphics[width=\linewidth]{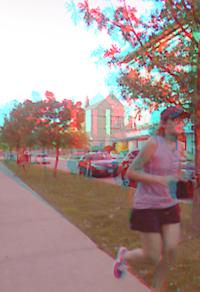}
\end{minipage}
\end{minipage}
\hfill
\begin{minipage}{.34\linewidth}
\centerline{{\bf Depth transfer (ours)}}
\centerline{215$\times$292}
\begin{minipage}{.5\linewidth}
\includegraphics[width=\linewidth]{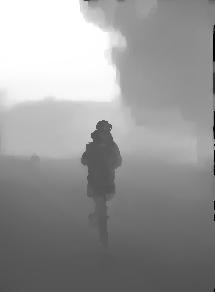}
\includegraphics[width=\linewidth]{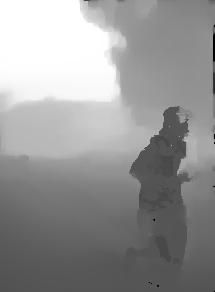}
\end{minipage}
\begin{minipage}{.445\linewidth}
\includegraphics[width=\linewidth]{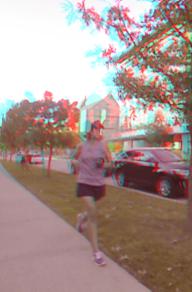}
\includegraphics[width=\linewidth]{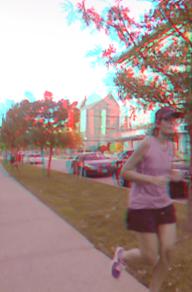}
\end{minipage}
\end{minipage}
\hfill
\vspace{-6mm}
\end{minipage}
\end{center} 
\caption{Comparison between our technique and the publicly available version of Make3D (\protect\url{http://make3d.cs.cornell.edu/code.html}). Make3D depth inference is trained to produce depths of resolution $55\times 305$ (bilinearly resampled for visualization), and we show results of our algorithm at the input native resolution. The anaglyph images are produced using the technique in Sec~\ref{sec:stereoest}. Depths displayed at same scale.
\vspace{2mm}
}
\label{fig:make3d_comparison}
\end{figure}

As further evaluation, a qualitative comparison between our technique and the publicly available version of Make3D is shown in Fig~\ref{fig:make3d_comparison}. Unlike Make3D, our technique is able to extract the depth of the runner throughout the sequence.

\begin{figure}
\begin{center}
\begin{minipage}{1.0\linewidth}
\begin{minipage}{0.49\linewidth}
\includegraphics[width=.195\linewidth]{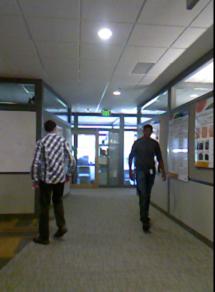}
\includegraphics[width=.195\linewidth]{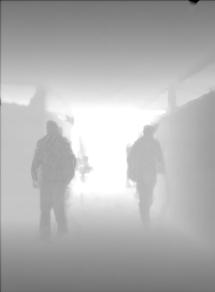}
\includegraphics[width=.195\linewidth]{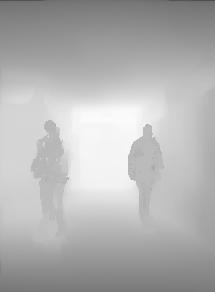}
\includegraphics[height=.265\linewidth]{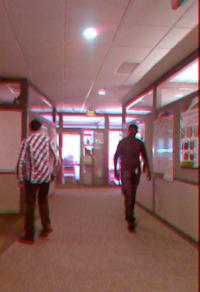}
\includegraphics[height=.265\linewidth]{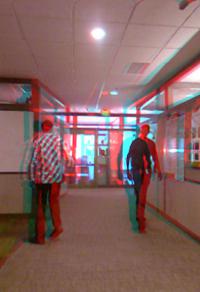}
\end{minipage}
\hfill
\rule[-7.5mm]{.15mm}{.13\linewidth}
\hfill
\begin{minipage}{.49\linewidth}
\includegraphics[width=.195\linewidth]{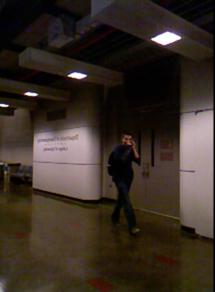}
\includegraphics[width=.195\linewidth]{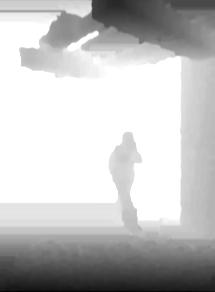}
\includegraphics[width=.195\linewidth]{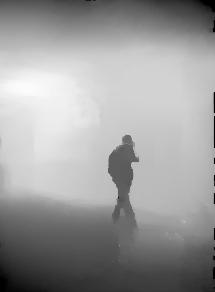}
\includegraphics[height=.265\linewidth, trim=0pt 0pt 0pt 0pt, clip=true]{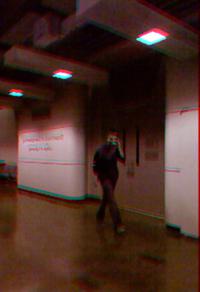}
\includegraphics[height=.265\linewidth, trim=0pt 0pt 0pt 0pt, clip=true]{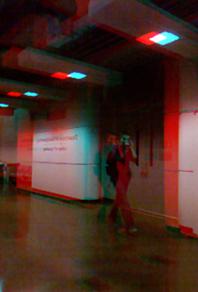}
\end{minipage}
\\
\rule{\linewidth}{.15mm}
\\
\begin{minipage}{.49\linewidth}
\includegraphics[width=.195\linewidth]{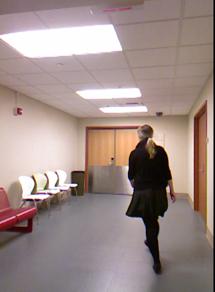}
\includegraphics[width=.195\linewidth]{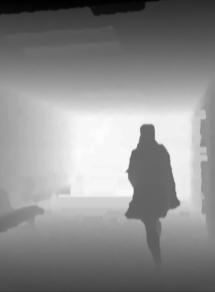}
\includegraphics[width=.195\linewidth]{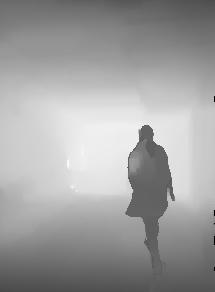}
\includegraphics[height=.265\linewidth]{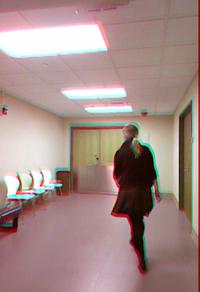}
\includegraphics[height=.265\linewidth]{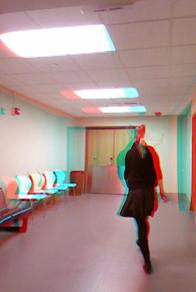}
\end{minipage}
\hfill
\rule[-7mm]{.15mm}{.13\linewidth}
\hfill
\begin{minipage}{.49\linewidth}
\includegraphics[width=.195\linewidth]{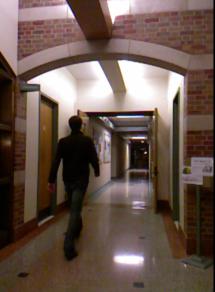}
\includegraphics[width=.195\linewidth]{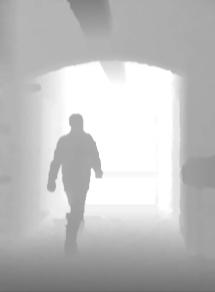}
\includegraphics[width=.195\linewidth]{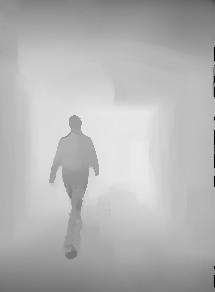}
\includegraphics[height=.265\linewidth, trim=0pt 0pt 0pt 0pt, clip=true]{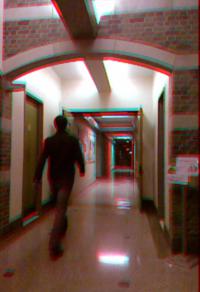}
\includegraphics[height=.265\linewidth, trim=0pt 0pt 4pt 0pt, clip=true]{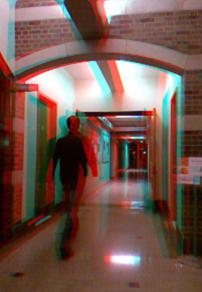}
\end{minipage}
\end{minipage}
\end{center}
\vspace{-5mm}
\caption{Video results obtained on test images for each building in our stereo-RGBD dataset (buildings 1-4, from left to right and top to bottom). For each result (from left to right): original photograph, ground truth depth, our inferred depth, ground truth anaglyph \protect\includegraphics[height=5pt]{fig/anaglyph.jpg} image, and our synthesized anaglyph image. Because the ground truth 3D images were recorded with a fixed interocular distance (roughly 5cm), we cannot control the amount of ``pop-out,'' and the 3D effect is subtle. However, this is a parameter we can set using our automatic approach to achieve a desired effect, which allows for an enhanced 3D experience. Note also that our algorithm can handle multiple moving objects (\emph{top}). Additional results are shown in the supplemental files.
\vspace{2mm}
}
\label{fig:indoor_results}
\end{figure}

\begin{figure}
\newcommand{\linespace}{\vspace{-1.em}}
\begin{minipage}{.325\linewidth}
\includegraphics[width=.49\linewidth]{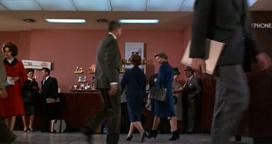}
\includegraphics[width=.49\linewidth]{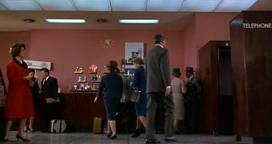}
\linespace\\
\includegraphics[width=.49\linewidth]{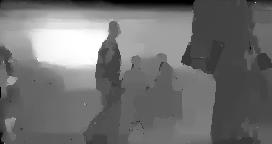}
\includegraphics[width=.49\linewidth]{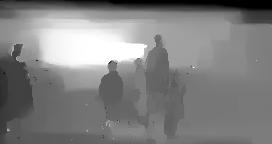}
\linespace\\
\includegraphics[width=.49\linewidth]{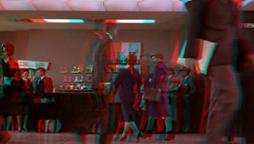}
\includegraphics[width=.49\linewidth]{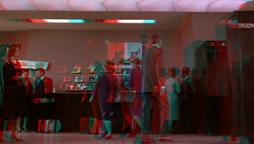}
\end{minipage}
\hfill
\begin{minipage}{.325\linewidth}
\includegraphics[width=.49\linewidth]{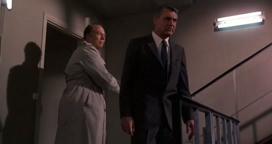}
\includegraphics[width=.49\linewidth]{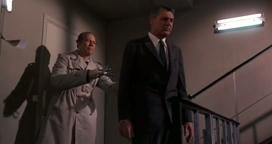}
\linespace\\
\includegraphics[width=.49\linewidth]{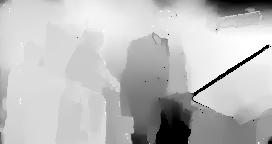}
\includegraphics[width=.49\linewidth]{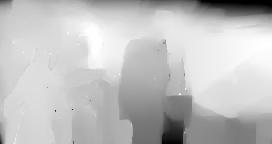}
\linespace\\
\includegraphics[width=.49\linewidth]{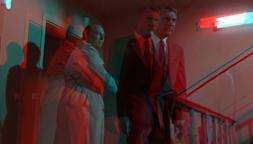}
\includegraphics[width=.49\linewidth]{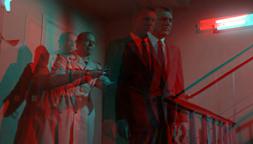}
\end{minipage}
\hfill
\begin{minipage}{.325\linewidth}
\includegraphics[width=.49\linewidth]{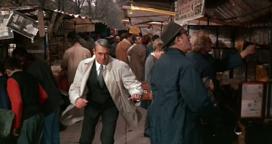}
\includegraphics[width=.49\linewidth]{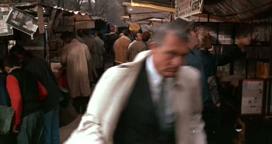}
\linespace\\
\includegraphics[width=.49\linewidth]{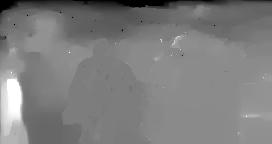}
\includegraphics[width=.49\linewidth]{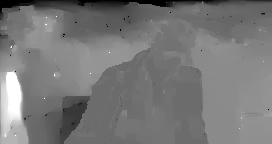}
\linespace\\
\includegraphics[width=.49\linewidth]{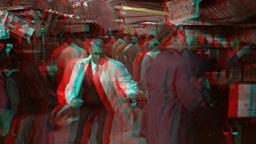}
\includegraphics[width=.49\linewidth]{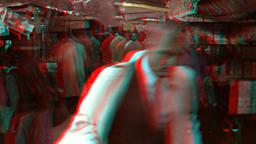}
\end{minipage}
\vspace{-2mm}
\caption{Several clips from the feature film {\it Charade}. Each result contains (from top to bottom): the original frames, estimated depth, and estimated anaglyph \protect\includegraphics[height=5pt]{fig/anaglyph.jpg} automatically generated by our algorithm. Some imperfections in depth are conveniently masked in the 3D image due to textureless or less salient regions.
}
\vspace{-8mm}
\label{fig:charade}
\end{figure}

Our technique works well for videos of many types scenes and video types (Figs ~\ref{fig:teaser},~\ref{fig:temporal_benefit},~\ref{fig:motionseg},~\ref{fig:rot_zoom_results},~\ref{fig:indoor_results}; more examples are in the supplemental files). We use the dataset we collected in Sec~\ref{sec:dataset} to validate our method for indoor scenes and videos (we know of no other existing methods/datasets to compare to). This dataset contains ground truth depth and stereo images for four different buildings (referred to as Buildings 1, 2, 3, and 4), and to demonstrate generalization, {\em we only use data from Building 1 for training}. We still generate results for Building 1 by holding each particular example out of the training set during inference.

We show quantitative results in Table~\ref{tab:results_kinect} and qualitative results in Fig~\ref{fig:indoor_results}. We calculate error using the same metrics as in our single image experiments, and to make these results comparable with Table~\ref{tab:results}, we globally rescale the ground truth and inferred depths to match the range of the Make3D database (roughly 1-81m). 
As expected, the results from Building 1 are the best, but our method still achieves reasonable errors for the other buildings as well.

Fig~\ref{fig:indoor_results} shows a result from each building in our dataset (top left is Building 1). As the quantitative results suggest, our algorithm performs very well for this building. In the remaining examples, we show results of videos captured in the other three buildings, all which contain vastly different colors, surfaces and structures from the Building 1. Notice that even for these images our algorithm works well, as evidenced by the estimated depth and 3D images.

Our algorithm also does not require video training data to produce video results. We can make use of static RGBD images (e.g. Make3D dataset) to train our algorithm for video input, and we show several outdoor video results in Figs~\ref{fig:teaser},~\ref{fig:temporal_benefit},~\ref{fig:motionseg} (more in the supplemental files). Even with static data from another location, our algorithm is usually able to infer accurate depth and stereo views.

Since we collected ground truth stereo images in our dataset, we also compare our synthesized right view with actual right view. We use peak signal-to-noise ratio ({\bf PSNR}) to measure the quality of the reconstructed views, as shown in Table~\ref{tab:results_kinect}. We could not acquire depth outdoors, and we use this metric to compare our outdoor and indoor results.

We also demonstrate that our algorithm may be suitable for feature films in Fig~\ref{fig:charade}. More diverse quantities of training are required to achieve commercial-quality conversion; however, even with a small amount of data, we can generate plausible depth maps and create convincing 3D sequences automatically. 

Recently, Youtube has released an automatic 2D-to-3D conversion tool, and we compared our method to theirs on several test sequences. Empirically, we noticed that the Youtube results have a much more subtle 3D effect. Both results are available online at \url{http://kevinkarsch.com/depthtransfer}.

Our algorithm takes roughly one minute per 640$\times$480 frame (on average) using a parallel implementation on a quad-core 3.2GHz processor.


\begin{figure}[t]
\begin{minipage}{0.48\linewidth}
\vspace{-3mm}
\begin{center}
\begin{minipage}{.218\linewidth}
\centerline{\bf\small Input}\vspace{0.5mm}
\includegraphics[width=\linewidth]{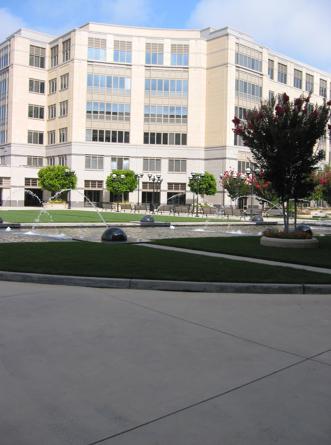}\vspace{0.5mm}\\
\includegraphics[width=\linewidth]{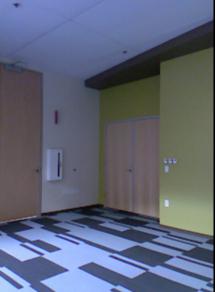}
\end{minipage}
\begin{minipage}{.22\linewidth}
\centerline{\bf\small Outdoor}\vspace{0.7mm}
\includegraphics[width=\linewidth]{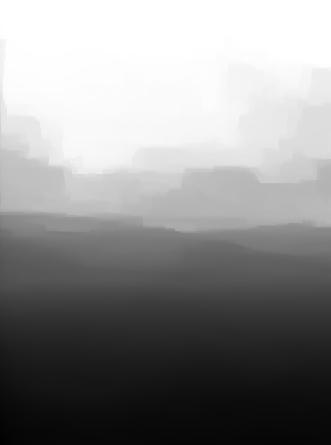}\vspace{0.5mm}\\
\includegraphics[width=\linewidth]{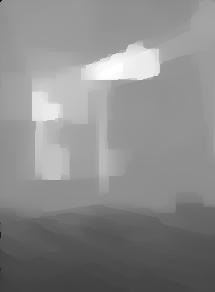}
\end{minipage}
\begin{minipage}{.22\linewidth}
\centerline{\bf\small Indoor}\vspace{0.5mm}
\includegraphics[width=\linewidth]{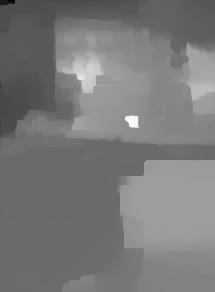}\vspace{0.5mm}\\
\includegraphics[width=\linewidth]{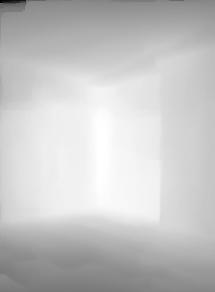}
\end{minipage}
\begin{minipage}{.22\linewidth}
\centerline{\bf\small All data}\vspace{0.7mm}
\includegraphics[width=\linewidth]{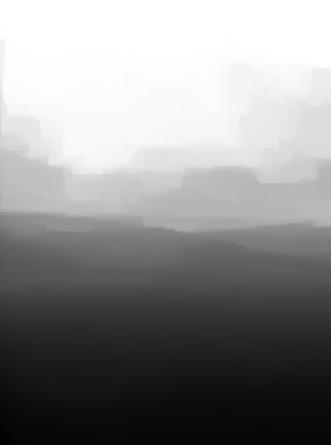}\vspace{0.5mm}\\
\includegraphics[width=\linewidth]{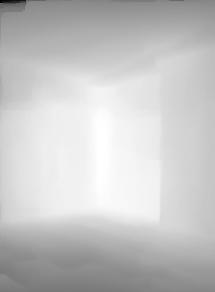}
\end{minipage}
\end{center}
\vspace{-5mm}
\caption{Effect of using different training data for indoor and outdoor images. While the results are best if the proper dataset is used, we also get good results even if we combine all the datasets.
}
\label{fig:train_example}\afterfigure
\end{minipage}
\hfill
\begin{minipage}{.48\linewidth}
\begin{center}
\includegraphics[width=.45\linewidth]{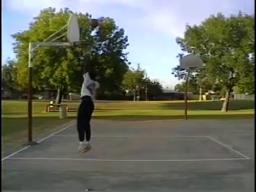}
\includegraphics[width=.45\linewidth]{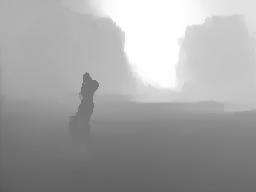}
\\
\includegraphics[width=.45\linewidth]{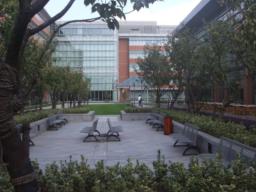}
\includegraphics[width=.45\linewidth]{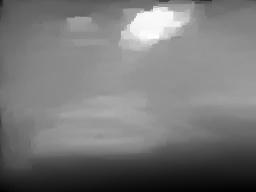}
\\\vspace{-5mm}
\end{center}
\caption{Example failure cases. Top row: thin or floating objects (pole and basketball) are ignored. Bottom row: input image is too different from training set. 
\vspace{-2mm}
}
\label{fig:basketball}
\end{minipage}
\end{figure}

\section{Discussion}
Our results show that our algorithm works for a large variety of indoor and outdoor sequences using a practical amount of training data. Note that our algorithm works for arbitrary videos, not just those with no parallax. However, videos with arbitrary camera motion and static scenes are best handled with techniques such as~\cite{Zhang:2011pami}. In Fig~\ref{fig:train_example}, we show that our algorithm requires some similar data in order to produce decent results (i.e., training with outdoor images for an indoor query is likely to fail). However, our algorithm can robustly handle large amounts of depth data with little degradation of output quality. The only issue is that more data requires more comparisons in the candidate search.

\begin{figure*}[t]

\begin{minipage}{.146\linewidth}
\centerline{\scriptsize Input}
\includegraphics[width=\linewidth]{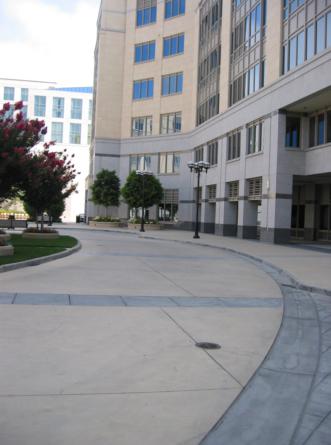}
\end{minipage}
\begin{minipage}{.535\linewidth}
\centerline{\scriptsize Candidate images and depths}
\includegraphics[width=.133\linewidth]{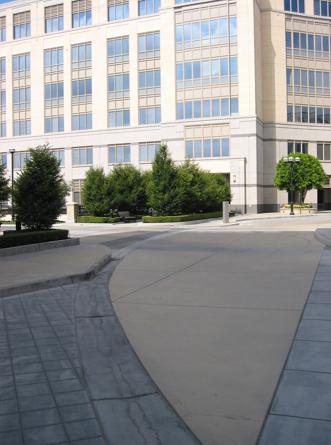}
\includegraphics[width=.133\linewidth]{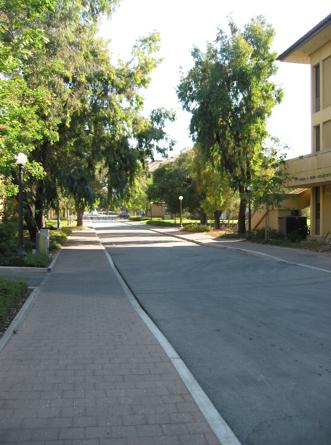}
\includegraphics[width=.133\linewidth]{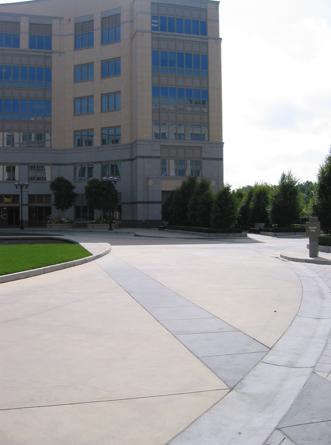}
\includegraphics[width=.133\linewidth]{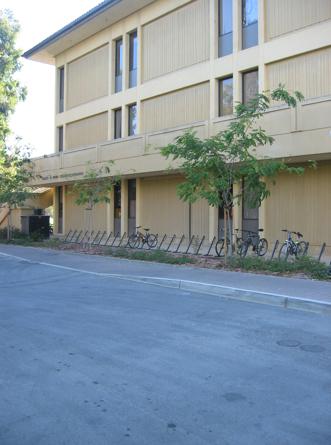}
\includegraphics[width=.133\linewidth]{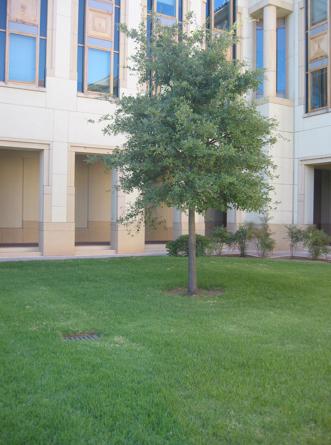}
\includegraphics[width=.133\linewidth]{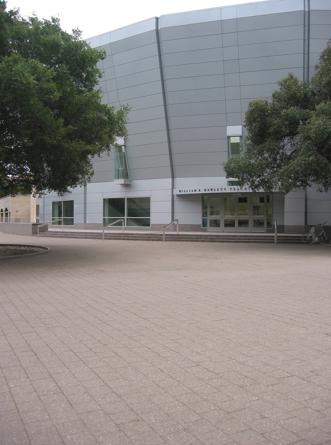}
\includegraphics[width=.133\linewidth]{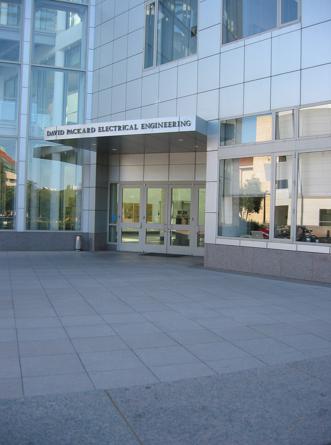}
\\
\includegraphics[width=.133\linewidth]{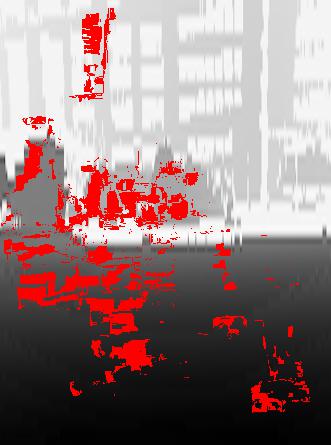}
\includegraphics[width=.133\linewidth]{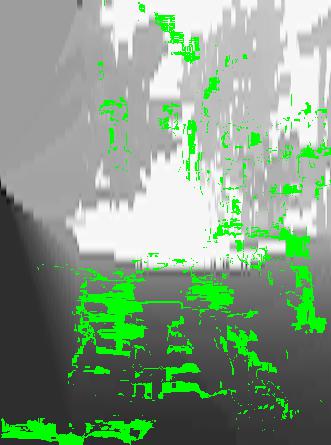}
\includegraphics[width=.133\linewidth]{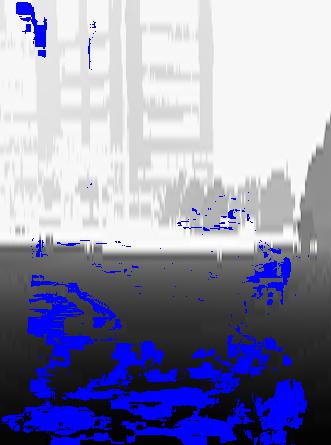}
\includegraphics[width=.133\linewidth]{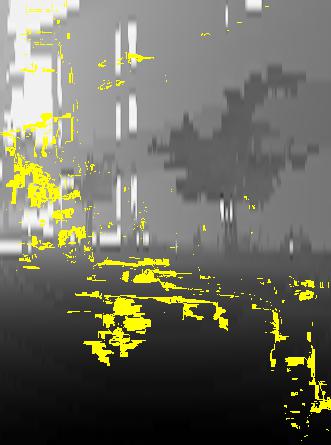}
\includegraphics[width=.133\linewidth]{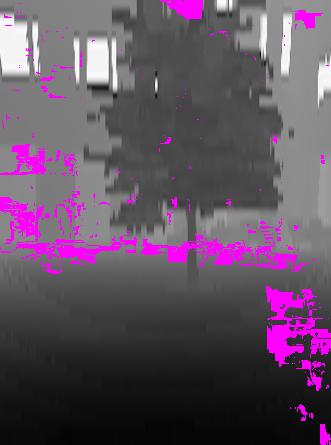}
\includegraphics[width=.133\linewidth]{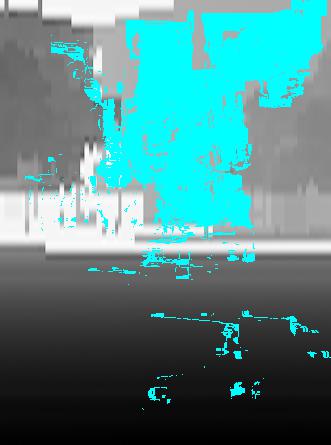}
\includegraphics[width=.133\linewidth]{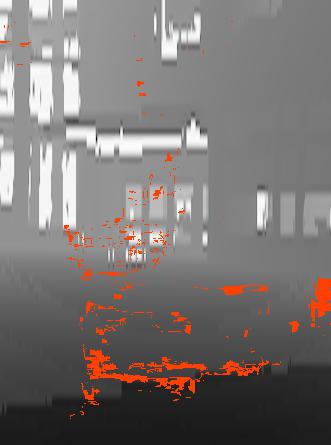}
\end{minipage}
\begin{minipage}{.146\linewidth}
\centerline{\scriptsize Contribution}
\vspace{.75mm}
\includegraphics[width=\linewidth]{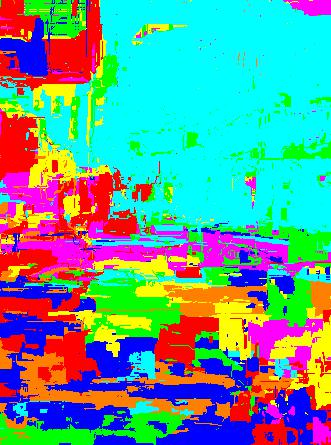}
\end{minipage}
\begin{minipage}{.146\linewidth}
\centerline{\scriptsize Inferred depth}
\includegraphics[width=\linewidth]{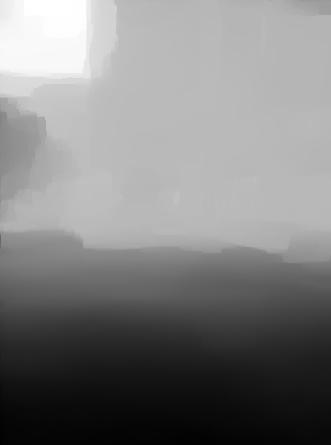}
\end{minipage}
\vspace{-3.5mm}
\caption{Candidate contribution for depth estimation. For an input image (\emph{left}), we find top-matching candidate RGBD images (\emph{middle}), and infer depth (\emph{right}) for the input using our technique. The contribution image is color-coded to show the sources; red pixels indicate that the left-most candidate influenced the inferred depth image the most, orange indicates contribution from the right-most candidate, etc.
\vspace{-1.5mm}
}
\label{fig:candidate_contribution}\vspace{-.5em}
\end{figure*}

This robustness is likely due to the features we use when determining candidate images as well as the design of our objective function. In Fig~\ref{fig:candidate_contribution}, we show an example query image, the candidates retrieved by our algorithm, and their contribution to the inferred depth. By matching GIST features, we detect candidates that contain features consistent with the query image, such as building facades, sky, shrubbery, and similar horizon location. Notice that the depth of the building facade in the input comes mostly from another similarly oriented building facade (teal), and the ground plane and shrubbery depth come almost solely from other candidates' ground and tree depths.

In some cases, our motion segmentation misses or falsely identifies moving pixels. This can result in inaccurate depth and 3D estimation, although our spatio-temporal regularization (Eqs.~\ref{eq:smooth},~\ref{eq:coherence}) helps to overcome this. Our algorithm also assumes that moving objects contact the ground, and thus may fail for airborne objects (see Fig~\ref{fig:basketball}).

Due to the serial nature of our method (depth estimation followed by view synthesis), our method is prone to propagating errors through the stages. For example, if an error is made during depth estimation, the result may be visually implausible. It would be ideal to use knowledge of how the synthesized views should look in order to correct issues in depth.

\section{Concluding Remarks}
We have demonstrated a fully automatic technique to estimate depths for videos. Our method is applicable in cases where other methods fail, such as those based on motion parallax and structure from motion, and works even for single images and dynamics scenes. Our depth estimation technique is novel in that we use a non-parametric approach, which gives qualitatively good results, and our single-image algorithm quantitatively outperforms existing methods. Using our technique, we also show how we can generate stereoscopic videos for 3D viewing from conventional 2D videos. Specifically, we show how to generate time-coherent, visually pleasing stereo sequences using our inferred depth maps. Our method is suitable as a good starting point for converting legacy 2D feature films into 3D. 

\section*{Acknowledgement}
We would like to thank Tom Blank for his critical help in creating our dual-Kinect data collection system. 

{\small
\bibliographystyle{splncs}
\bibliography{eccv12-autostereo}
}

\end{document}